\documentclass[10pt,conference]{IEEEtran}

\IEEEoverridecommandlockouts
\usepackage{graphicx, lipsum}
\usepackage{hyperref}       % hyperlinks
\usepackage{url}            % simple URL typesetting
\usepackage{booktabs}       % professional-quality tables
\usepackage{amsfonts}       % blackboard math symbols
\usepackage{nicefrac}       % compact symbols for 1/2, etc.
\usepackage{microtype}      % microtypography
\usepackage[table]{xcolor}  % colors
\usepackage{comment}
\usepackage{gensymb}
\usepackage{placeins}
\usepackage{amsmath}        % ← Move BEFORE wasysym
\usepackage{wasysym}        % ← Now loads after amsmath
\usepackage{graphicx}
\usepackage{xspace}
\usepackage{algorithm}
\usepackage{algpseudocode}
\usepackage{enumitem} % Essential for [leftmargin=*]
\usepackage[caption=false,font=footnotesize]{subfig}
\usepackage{xcolor}

\usepackage{cite}
\usepackage{scalerel} %for orcid
\usepackage{tikz} % figure package
\usepackage{stfloats}
\usetikzlibrary{svg.path} %for orcid

\definecolor{poscolor}{HTML}{3D405B}
\definecolor{orcidlogocol}{HTML}{A6CE39}
\tikzset{
  orcidlogo/.pic={
    \fill[orcidlogocol] svg{M256,128c0,70.7-57.3,128-128,128C57.3,256,0,198.7,0,128C0,57.3,57.3,0,128,0C198.7,0,256,57.3,256,128z};
    \fill[white] svg{M86.3,186.2H70.9V79.1h15.4v48.4V186.2z}
                 svg{M108.9,79.1h41.6c39.6,0,57,28.3,57,53.6c0,27.5-21.5,53.6-56.8,53.6h-41.8V79.1z M124.3,172.4h24.5c34.9,0,42.9-26.5,42.9-39.7c0-21.5-13.7-39.7-43.7-39.7h-23.7V172.4z}
                 svg{M88.7,56.8c0,5.5-4.5,10.1-10.1,10.1c-5.6,0-10.1-4.6-10.1-10.1c0-5.6,4.5-10.1,10.1-10.1C84.2,46.7,88.7,51.3,88.7,56.8z};
  }
}

\newcommand\orcidicon[1]{\href{https://orcid.org/#1}{\mbox{\scalerel*{
\begin{tikzpicture}[yscale=-1,transform shape]
\pic{orcidlogo};
\end{tikzpicture}
}{|}}}}

\begin{document}

\newcommand{\algname}{\textsc{FARS}\xspace}

\title{Fuzzy Logic Theory-based Adaptive Reward Shaping for Robust Reinforcement Learning (\algname)}

\author{Hürkan Şahin\textsuperscript{\orcidicon{0009-0008-7920-5872}}, 
Van Huyen Dang\textsuperscript{\orcidicon{0009-0006-2328-2153}}, 
Erdi Sayar\textsuperscript{\orcidicon{0000-0002-5944-7291}}, 
Alper Yegenoglu\textsuperscript{\orcidicon{0000-0001-8869-215X}}, and 
Erdal Kayacan\textsuperscript{\orcidicon{0000-0002-7143-8777}}
\thanks{*This work was partially supported by the Horizon Europe Grant Agreement No.  101136056 and No. 101119774.}% <-this % stops a space
\thanks{Hürkan Şahin, Van Huyen Dang, Erdi Sayar, Alper Yegenoglu, and Erdal Kayacan are with the Automatic Control Group (RAT), Paderborn University, 33098 Paderborn, Germany
        {\tt\footnotesize \{hursah, van.huyen.dang, erdi.sayar, alper.yegenoglu, erdal.kayacan\}@upb.de}}
        }%  
\maketitle

\begin{abstract}
Reinforcement learning (RL) often struggles in real-world tasks with high-dimensional state spaces and long horizons, where sparse or fixed rewards severely slow down exploration and cause agents to get trapped in local optima. This paper presents a fuzzy-logic-based reward shaping method that integrates human intuition into RL reward design.
By encoding expert knowledge into adaptive, interpretable terms, fuzzy rules promote stable learning and reduce sensitivity to hyperparameters. The proposed method leverages these
properties to adapt reward contributions based on the agent’s state, enabling smoother transitions between fast motion and precise control in challenging navigation tasks. The extensive
simulation results on autonomous drone racing benchmarks show stable learning behavior and consistent task performance across scenarios of increasing difficulty. The proposed method achieves
faster convergence and reduced performance variability across training seeds in more challenging environments, with success rates improving by up to approximately 5\% compared to non-fuzzy
reward formulations.
\end{abstract}

\IEEEpeerreviewmaketitle

\section{Introduction}
\label{sec:intro}
Deep reinforcement learning (RL) has delivered strong performance across diverse domains by learning policies directly from high-dimensional sensory inputs, such as raw pixels or sensor data. Key examples include  robotic manipulation \cite{10.1145/3638529.3654045, 10610910} and autonomous drone navigation \cite{11152316,11007627}. Despite these successes, RL performance in complex tasks remains highly sensitive to the design of the reward function, as sparse or poorly balanced rewards can hinder exploration and destabilize training. While reward shaping \cite{ng1999policy} offers denser feedback, conventional approaches rely on manually tuned reward combinations that are often brittle and sensitive to task variations. 

Fuzzy logic has been widely adopted in unmanned aerial vehicles (UAVs) ~\cite{7580570,8304792,8809217, SARABAKHA2019105495, CAMCI20181,7737744} to handle uncertainties and rapidly changing operating conditions. Recent studies have explored integrating fuzzy logic into RL frameworks to improve robustness. A control framework based on proximal policy optimization (PPO) \cite{schulman2017proximal} is introduced in \cite{lyapunov_fuzzy}. The authors incorporate a Lyapunov-inspired fuzzy reward and an adjustable policy learning rate to improve training stability.  A PPO-based actor–critic framework is presented in \cite{Bingolsafe}, where manually tuned classical rewards are replaced with a fuzzy logic–based reward formulation. This results in robust navigation performance in both static and dynamic environments. These works collectively indicate that fuzzy logic can provide structured, interpretable, and stabilizing reward signals for RL.
\begin{figure}[t!]
    \centering
    \includegraphics[width=0.95\linewidth]{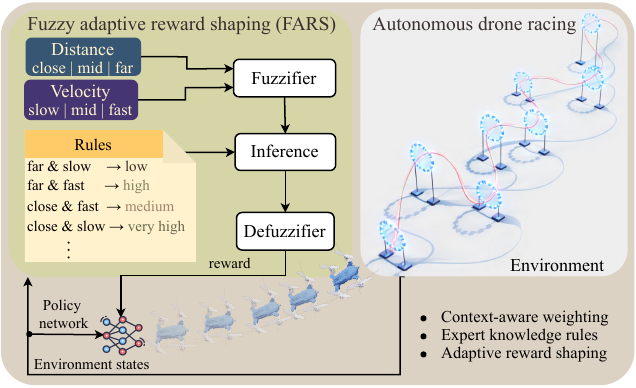}
    \vspace{-3mm} %please dont use manual reduce
\caption{Schematic overview of the fuzzy adaptive reward shaping (FARS) method. Whereas conventional RL design typically relies on manually tuned and crisp input terms, FARS introduces a fuzzy logic-based method that adaptively shapes reward as a function of distance and velocity using human-interpretable linguistic rules. The fuzzy reward is integrated into the RL objective to provide context-aware feedback, enabling faster convergence, reduced variance, and improved stability.  
%The method is demonstrated on autonomous drone racing tasks, where high-speed navigation through sequential gates benefits from adaptive and interpretable reward shaping.
}
\vspace{-2mm} 
    \label{fig:abstract}
\end{figure}

Autonomy in fast and agile flight has emerged as a powerful benchmark in robotics, as it simultaneously challenges perception, planning, and control.  Furthermore, fast and agile flight is crucial for enabling autonomous drones to tackle time-critical real-world tasks; e.g. rapidly searching cluttered disaster zones for survivors. However, at high speeds with rapid rotations, nonlinear system dynamics are excited, providing a demanding testbed for advanced navigation methods~\cite{faessler2018differential, robotics11050109, 10649903, 9894655, 9838538, PHAM2022371, 9636207, 9206943}.  As a challenging benchmark for reward design, autonomous drone racing requires an aerial robot to navigate at high speed through a sequence of gates  under long-horizon and real-time constraints. Recent works have explored deep RL for drone racing, primarily focusing on generalization and robustness. An adaptive environment-shaping framework is proposed in \cite{wang2026}, where progressively challenging training environments are generated by a secondary RL agent, enabling a single racing policy to generalize to unseen and dynamic tracks.
A two-phase, curriculum-based RL framework is introduced in \cite{yu2025master}, in which aggressive flight and collision avoidance are balanced to achieve robust vision-based drone racing performance. 

Recent work on fuzzy-logic–based RL has shown improved stability and interpretability through structured reward formulations. This work explores reward-level abstraction for autonomous drone racing by encoding human-inspired heuristics—such as maintaining high speed over long distances and decelerating near gates—directly into the reward function using fuzzy linguistic rules, as illustrated in Fig.~\ref{fig:abstract}. This approach yields a simple, interpretable reward formulation that generalizes across racing scenarios without task-specific reward engineering.

The main contributions of this paper are as follows:
\begin{itemize}
\item We propose an interpretable fuzzy-logic reward shaping method called FARS. It automatically balances fast movement when far from gates and slow, precise movement when close to gates. This results in smoother and more intelligent rewards than traditional non-fuzzy (crisp) shaping methods.
% \item We test and compare the methods carefully on drone racing tasks of increasing difficulty (Easy, Medium, and Hard zigzag courses) in IsaacLab~\cite{mittal2025isaaclab}. 
\item We compare our proposed fuzzy logic-based reward shaping method using Mamdani and Sugeno defuzzification against a well-known potential field-based method \cite{ng1999policy} on drone racing tasks of increasing difficulty (Easy, Medium, and Hard zigzag courses). We analyze and discuss the main differences in smoothness, training stability, and final performance. 
\item We show that fuzzy rewards help the drone switch smoothly between fast flying and careful slowing in long and difficult drone racing tasks. By using velocity and distance in a natural way, fuzzy rewards lead to faster learning and higher success rates compared to non-fuzzy methods.
\end{itemize}

The remainder of this paper is organized as follows: Section~\ref{sec:intro} introduces the problem and reviews related work. Section~\ref{sec:methodology} presents the proposed fuzzy logic–based reward shaping method. Section~\ref{sec:experiments} describes the experimental setup, while Section~\ref{sec:results} reports and discusses the experimental results. Finally, Section~\ref{sec:conclusion} concludes the paper and outlines directions for future work.

\vspace{-0.2cm}
\section{Methodology}
\label{sec:methodology}
To demonstrate the effectiveness of the proposed fuzzy-based reward shaping method, we employ a deep RL framework to solve a drone racing task, as illustrated in Fig.~\ref{fig:zigzag_scenario}. 

This task is modeled as a Markov decision process defined by the tuple $(S, A, P, r, \gamma)$, where $S$ and $A$ denote the state and action spaces, $P$ is the transition matrix, $r$ is the reward function, and $\gamma \in [0,1)$ is the discount factor~\cite{puterman1990markov}.
Accordingly, the objective is to learn a policy $\pi_\theta$ that maximizes the expected discounted return:
\begin{equation}
\max_{\theta}\; \mathbb{E}\!\left[\sum_{t=0}^{\infty} \gamma^t\, r_t \right].
\label{eq:policy}
\end{equation}
Since reward design plays a crucial role in learning efficiency, we next present the proposed fuzzy logic–based reward shaping approach and compare it with a commonly used potential-based reward shaping method \cite{ng1999policy}.
\subsection{Potential field-based reward shaping (PFBRS) for drone racing}
In principle, a reward function is designed to guide the agent's learning process as it navigates a sequence of gates and hovers at the end point. 
We define a reward function, including multiple terms: 
% a gate-passage reward ($r^\text{gate}$), a final-goal reward ($r^\text{final}$), a time penalty ($r^\text{time}$),($r^\text{dist}$) distance-to-goal reward, and a collision/termination penalty ($r^\text{die}$).
\subsubsection{Gate-passed reward}
The gate-passed reward $r^\text{gate}$ increases cumulatively with the number of gates successfully passed, providing positive reinforcement for sequential progress.
\begin{equation}
r^{\text{gate}} =
\begin{cases}
c_1, & \text{if the agent passes a gate at time } t,\\
0,  & \text{otherwise.}
\end{cases}
\end{equation}
\subsubsection{Final goal hovering reward}
After the agent passes the final gate, the task switches to a goal-hovering phase. A dense distance-based reward encourages precise positioning at the final goal. Let $p_t$ and $p_{\mathrm{goal}}$ denote the agent and goal positions, respectively. The reward is defined as
\begin{equation}
r^{\mathrm{final}} =
c_2 \exp\left(- \lVert p_t - p_{\mathrm{goal}} \rVert_2 \right)\Delta t
\end{equation}
%\subsubsection{Time penalty}
%The time penalty $r^\text{time}$ is a small negative reward applied at every timestep, encouraging the drone to complete the task as quickly as possible.
%\begin{equation}
%r^{\text{time}} = - \Delta t .
%\end{equation}
%\color{black}
\begin{figure}[t!]
\centering
\includegraphics[width=0.975\linewidth]{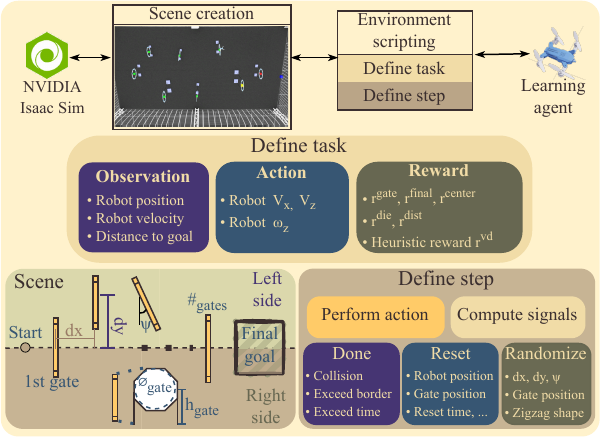}
\vspace{-2mm}
\caption{Overview of our proposed training environment  and the task design in IsaacLab~\cite{mittal2025isaaclab}. The top row illustrates the task formulation, including observations, continuous actions, and the composite reward structure. The middle row shows the interaction between Isaac Sim--based scene generation, environment scripting, and the learning agent. The bottom row depicts the zigzag gate racing scenario with randomized gate positions and orientations, ending in a final goal region.}

\label{fig:zigzag_scenario}
\end{figure}
\subsubsection{Distance-to-gate reward}
A dense shaping reward is applied based on the Euclidean distance to the
currently active gate (or the final goal after the last gate):
\begin{equation}
r^\text{dist} = \left\| p_t - p_{\mathrm{goal}} \right\|_2 \Delta t
\end{equation}
\subsubsection{Gate alignment reward}
The gate center alignment reward $r^{\mathrm{center}}$, which encourages the agent to align with the gate center for easier gate passing.
\begin{equation}
    r^{\text{center}} =
    \exp\!\left(- \left| \mathbf{n} \cdot \mathbf{v}_{\text{agent}} \right|\right)
\end{equation}
where $\left| \mathbf{n} \cdot \mathbf{v}_{\text{agent}} \right|$ describes the alignment between the agent-to-gate moving direction $\mathbf{v}_{\text{agent}}$ and the direction of the normal vector crossing the gate center $\mathbf{n}$.
\subsubsection{Collision penalty}
The collision penalty $r^\text{die}$ is a large negative reward applied if the drone collides with gates.
\begin{equation}
r^{\text{die}} =
\begin{cases}
-c_3, & \text{if a collision occurs at time } t,\\
0,   & \text{otherwise.}
\end{cases}
\end{equation}
% \color{blue}
% To improve learning stability, we apply potential-based reward shaping to the dense terms. Given a potential function $\Phi:S\rightarrow\mathbb{R}$, the shaping reward is defined as
% \begin{equation}
% F(s_t, s_{t+1}) = \gamma\,\Phi(s_{t+1}) - \Phi(s_t).
% \label{eq:shaping}
% \end{equation}

\color{black}
We apply potential-based shaping to the terms $r^{\mathrm{dist}}$ and $r^{\mathrm{center}}$
to provide intermediate feedback and prevent training divergence. In general, the potential field function is constructed as 
\begin{equation}
r_{\mathrm{aux}} = \gamma \times r (s_{t}) - r(s_{t-1})
\label{eq:shaping}
\end{equation}
where $ r(s_{t-1})$ is the previous reward value, and $r(s_{t})$ is the current value. Hence, the overall reward function with a potential-field-based approach is defined as
\begin{equation}\label{eq:pbrs}
    r_t = r^\text{gate} + r^\text{final} + r_{\mathrm{aux}}^\text{dist}+ r_{\mathrm{aux}}^{\mathrm{center}} + r^\text{die} 
\end{equation}
The remaining terms are retained unchanged, as they represent single-event rewards or penalties. The selection of hyperparameters, $c_1, c_2, c_3$, is commonly performed via trial and error and is highly reliant on the practitioner’s experience with reward shaping.
\subsection{Reward shaping enhancement with fuzzy-logic theory}
A fuzzy-logic–based reward formulation is introduced to explicitly address the trade-off between approach speed and precision during gate traversal. Specifically, a velocity–distance reward $r^{\mathrm{vd}}$ is designed by defining intuitive linguistic rules over distance and velocity states, which enables smooth interpolation of the reward signal to promote the desired behaviours.

\begin{figure}[t]
  \centering

  % ===== Row 1 =====
  \begin{minipage}[t]{0.24\textwidth}\centering
    \subfloat[Mamdani $r^{\mathrm{vd}}$]{%
      \includegraphics[width=\linewidth]{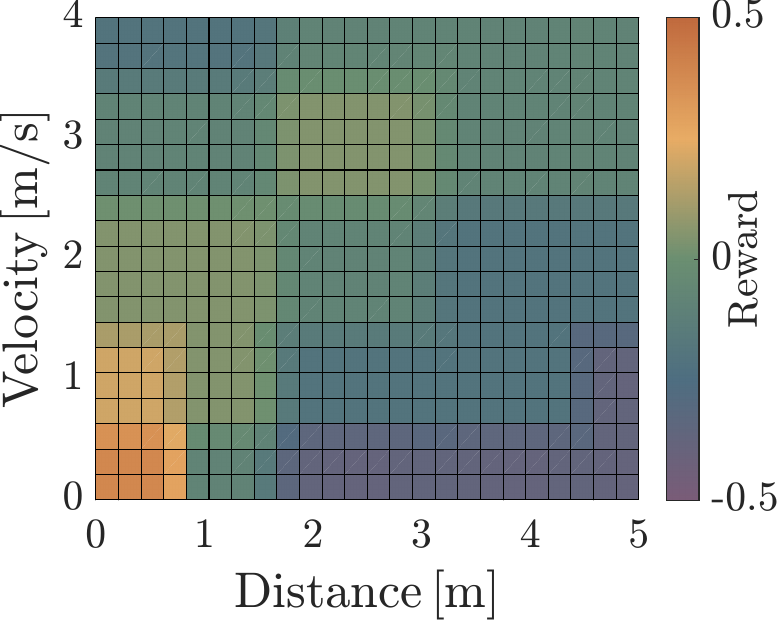}%
      \label{fig:mamdani}}
  \end{minipage}\hfill
  \begin{minipage}[t]{0.24\textwidth}\centering
    \subfloat[Sugeno $r^{\mathrm{vd}}$]{%
      \includegraphics[width=\linewidth]{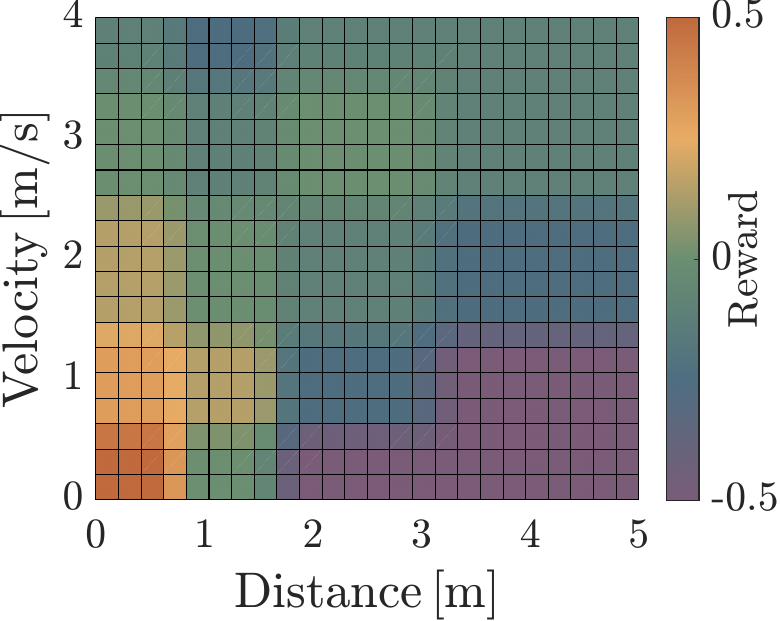}%
      \label{fig:sugeno}}
  \end{minipage}\hfill
  %\begin{minipage}[t]{0.16\textwidth}\centering
  %  \subfloat[Non-Fuzzy $r_t^{\mathrm{vd}}$]{%
  %    \includegraphics[width=\linewidth]{figures/fuzzy_analytic_surface.pdf}%
  %    \label{fig:non-fuzzy}}
  %\end{minipage}

\caption{Velocity--distance reward surfaces $r^{vd}$ derived from (a) Mamdani and (b) Sugeno fuzzy inference systems. The surfaces provide a continuous representation of the fuzzy rule base, mapping distance--velocity inputs to reward outputs. Higher reward values correspond to higher velocities at larger distances and lower velocities near the target.}
    \label{fig:fuzzy_reward_surfaces}
\end{figure}

Within the fuzzy-based formulation, the classical distance-based auxiliary reward $r_{\mathrm{aux}}^{\mathrm{dist}}$ is replaced by the fuzzy velocity–distance reward, while all other reward components remain unchanged. Consequently, the total reward in the fuzzy-based approach is defined as follows:
\begin{equation}\label{eq:fuzzy}
    r_t =
    r^{\text{gate}}
    + r^{\text{final}}
    + r^{\mathrm{vd}}
    + r_{\mathrm{aux}}^{\text{center}}
    + r^{\text{die}}.
\end{equation}
The fuzzy-logic-based reward depends on two inputs: the distance to the currently active gate and the magnitude of the drone's linear velocity. The velocity-distance reward is defined using two alternative formulations, each encoding the same underlying heuristic rather than introducing distinct behavioural objectives:
(i) Mamdani fuzzy inference ~\cite{mamdani1974application}, (ii) Sugeno fuzzy inference~\cite{sugeno1985fuzzy}, with rule-based surfaces shown in Fig.~\ref{fig:mamdani} and Fig.~\ref{fig:sugeno},
In all cases, the same distance and velocity normalization, membership functions, and reward logic are employed.
The intuition behind the fuzzy logic is as follows:
$$
\text{If } v \text{ is } A_i \text{ and } d \text{ is } B_i, \text{ then } z \text{ is } C_i,
$$
for example: if \textit{"the drone is far from the target gate"}, and \textit{"it flies at a high velocity"}, then this behavior is encouraged to promote rapid progress. Similarly, if \textit{"the drone approaches the gate"}, and \textit{"its velocity is slow"}, then this behavior is also rewarded due to safe, stable, and centered traversal. 

At comparable velocities, closer proximity to the gate yields a higher reward than maintaining a greater distance, thereby explicitly encouraging rapid convergence toward the gate rather than slow movement at greater distances.  This design is essential to prevent conservative behaviours, such as premature speed reduction without meaningful progress. Fig.~\ref{fig:fuzzy_reward_surfaces} illustrates how this principle is realised by the three velocity--distance reward formulations. Although the Mamdani and Sugeno variants differ in terms of smoothness and representation, both maintain the same qualitative structure. The Mamdani formulation produces a smooth, continuous reward surface, while the Sugeno formulation displays piecewise regions due to constant rule outputs.

\section{Experiments}
\label{sec:experiments}
%\subsection{Test Scenarios}

\begin{figure}[b]
\centering
\begin{minipage}{0.31\linewidth}
    \centering
    \includegraphics[width=\linewidth]{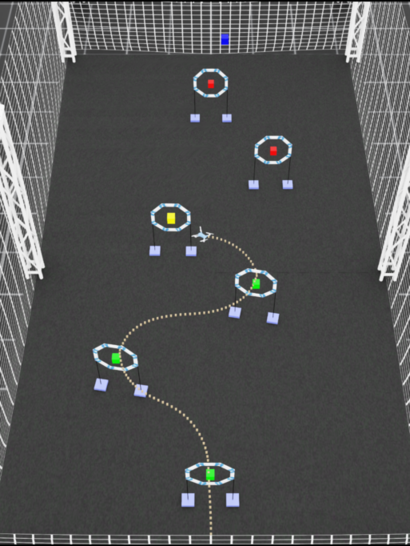}
\end{minipage}
\hfill
\begin{minipage}{0.31\linewidth}
    \centering
    \includegraphics[width=\linewidth]{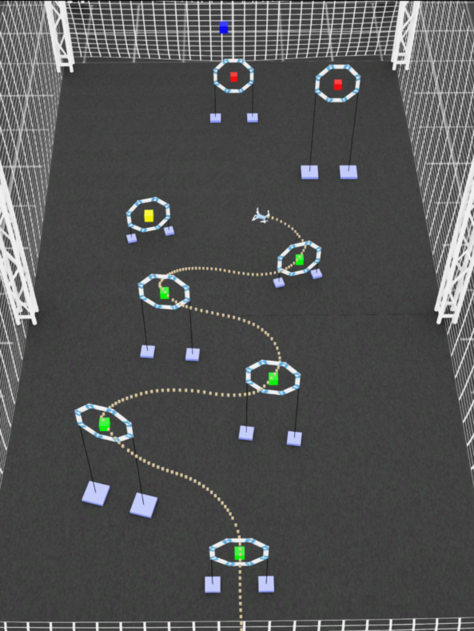}
\end{minipage}
\hfill
\begin{minipage}{0.31\linewidth}
    \centering
    \includegraphics[width=\linewidth]{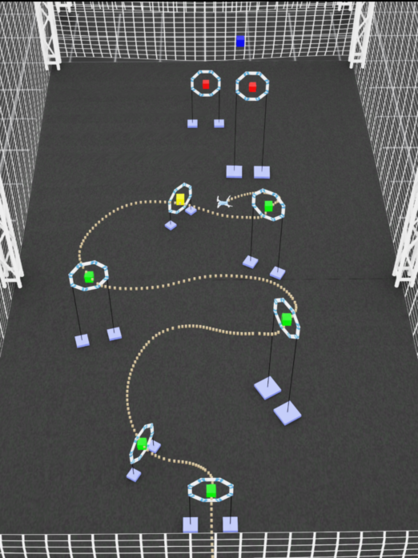}
\end{minipage}

\caption{Randomly generated zigzag environments in Isaac Sim for the Easy, Medium, and Hard scenarios (left to right).
The environments are confined within predefined spatial boundaries.
The dotted curve shows the drone trajectory.
Green markers denote passed gates, the yellow marker indicates the active target gate, and red markers represent inactive gates.
Each episode is limited to $15\mathrm{s}$ and terminates immediately upon collision.}
\label{fig:zigzag_env}
\end{figure}
The navigation policy is evaluated on a zigzag gate circuit designed to enforce alternating lateral maneuvers while maintaining forward motion, as illustrated in Fig.~\ref{fig:zigzag_scenario} scene section. Gates are arranged sequentially along the $x$-axis, with the first and last gates centered on the $y$-axis. Intermediate gates alternate strictly between left ($y>0$) and right ($y<0$) placements, forming a zigzag trajectory. Lateral offsets are sampled from $\ 0.5\ \mathrm{m},\ 0.75\ \mathrm{m},\ 1.0\ \mathrm{m},\ 1.25\ \mathrm{m},\ 1.5\ \mathrm{m},\ 1.75\ \mathrm{m}\ $ with a uniform perturbation of up to $0.1$m.

The Easy scenario focuses on basic lateral alternation and forward motion. The course consists of $6$ gates with fixed height ($1.0\ \mathrm{m}$) and diameter ($0.60\ \mathrm{m}$). The longitudinal distance between consecutive gates is sampled from $1.5\ \mathrm{m},  1.75\ \mathrm{m}, 2.0\ \mathrm{m} $. Small yaw perturbations are applied to the gates, uniformly sampled from $[-10^\circ,\,10^\circ]$. This configuration provides generous clearance and limited variability, serving as a baseline for navigation performance.

The Medium scenario increases complexity by introducing tighter longitudinal spacing and vertical variability. The number of gates is increased to eight, while the distance between gates is reduced to $ 1.0\ \mathrm{m},\ 1.25\ \mathrm{m}, 1.5\ \mathrm{m} $. The gate diameters remain unchanged, but the gate heights are $0.5\ \mathrm{m},\ 1.0\ \mathrm{m},\ 1.5\ \mathrm{m},\ 2.0\ \mathrm{m}$. This scenario requires the policy to handle increased vertical variation. The Hard scenario further increases difficulty by reducing the gate diameter to $0.45 \mathrm{m}$ and introducing larger yaw perturbations sampled from $[-60^\circ,60^\circ]$. Consequently, the Hard scenario imposes substantially greater demands on precise trajectory control, accurate yaw alignment, and robust perception during rapid alternating maneuvers. Example environments for each difficulty level are visualized in Fig.~\ref{fig:zigzag_env}. In simulation, the UAV is modeled as an X-configuration quadrotor with 5\,inch propellers, having a propeller tip--to--tip footprint of \(0.325\,\mathrm{m} \times 0.325\,\mathrm{m}\) and a height of \(0.08\,\mathrm{m}\).

\begin{figure*}[t]
\centering
\setlength{\tabcolsep}{3pt}
\renewcommand{\arraystretch}{1.1}

\begin{tabular}{c c c c c}
& \multicolumn{3}{c}{\small Training} 
& \multicolumn{1}{c}{\small Evaluation} \\
\cmidrule(lr){2-4} \cmidrule(lr){5-5}
\vspace{-2em} \\
% --- Row 1: Easy ---
\raisebox{1.2cm}{\rotatebox{90}{\small Easy}} \label{row:easy} &
\subfloat[]{\includegraphics[width=0.225\textwidth]{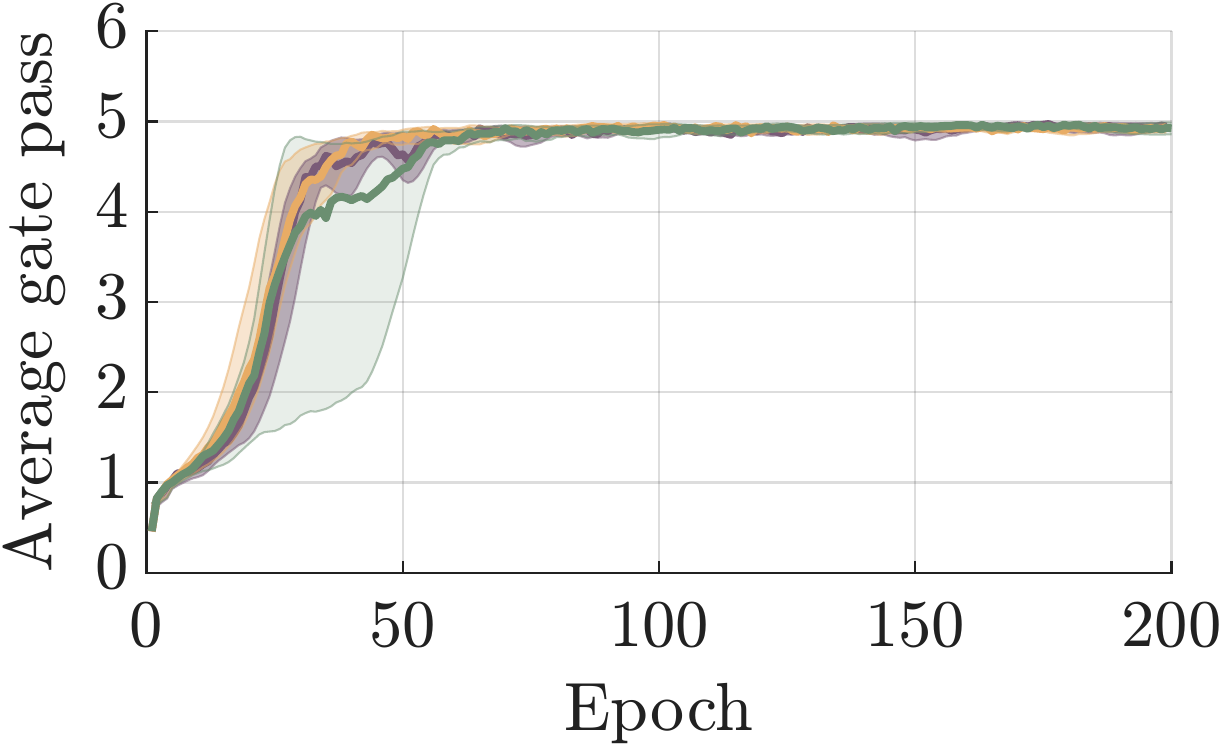}\label{fig:easy_gate_passed}} &
\subfloat[]{\includegraphics[width=0.23\textwidth]{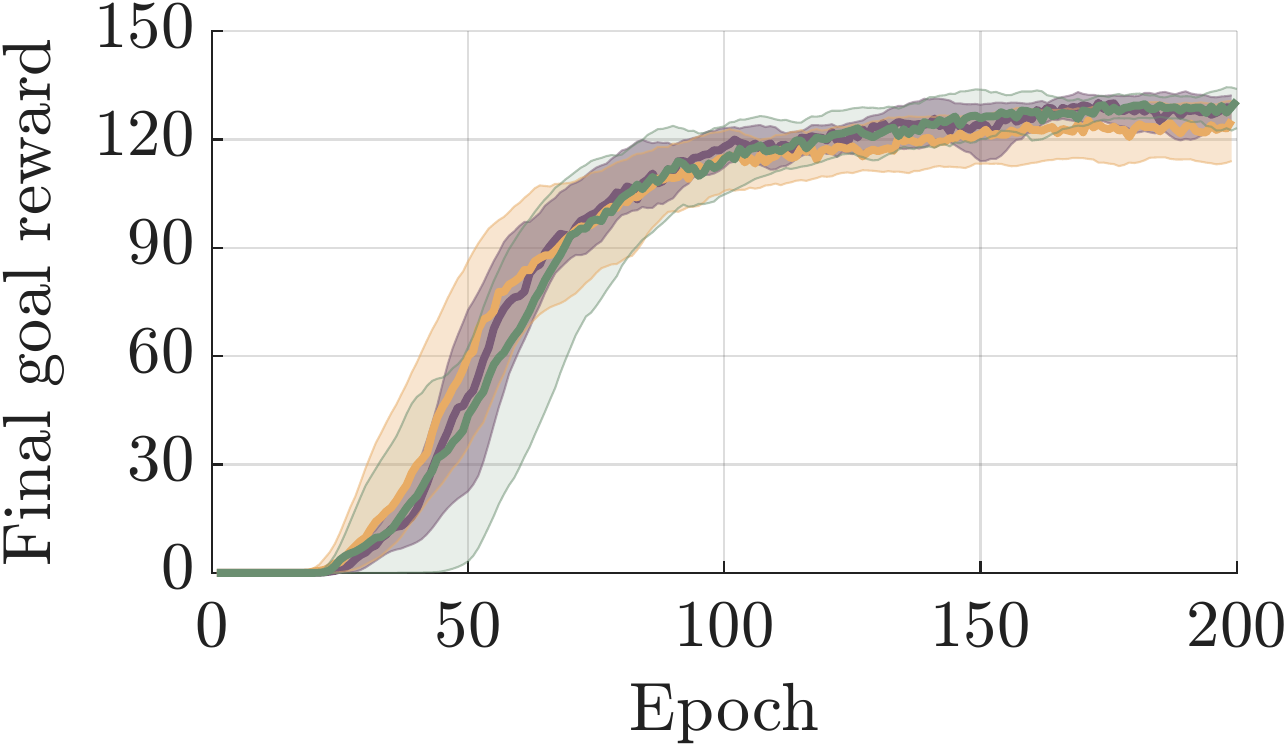}\label{fig:easy_goal_hover}} &
\subfloat[]{\includegraphics[width=0.225\textwidth]{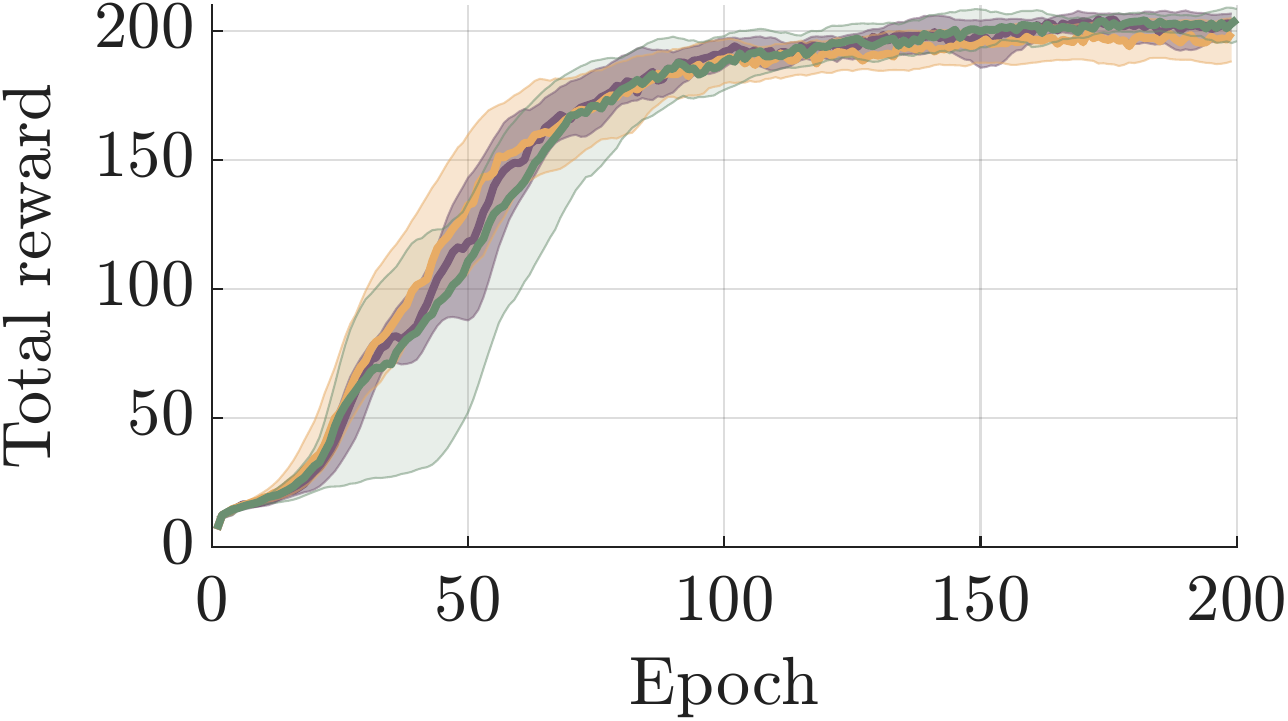}\label{fig:easy_total_reward}} &
\subfloat[]{\includegraphics[width=0.22\textwidth]{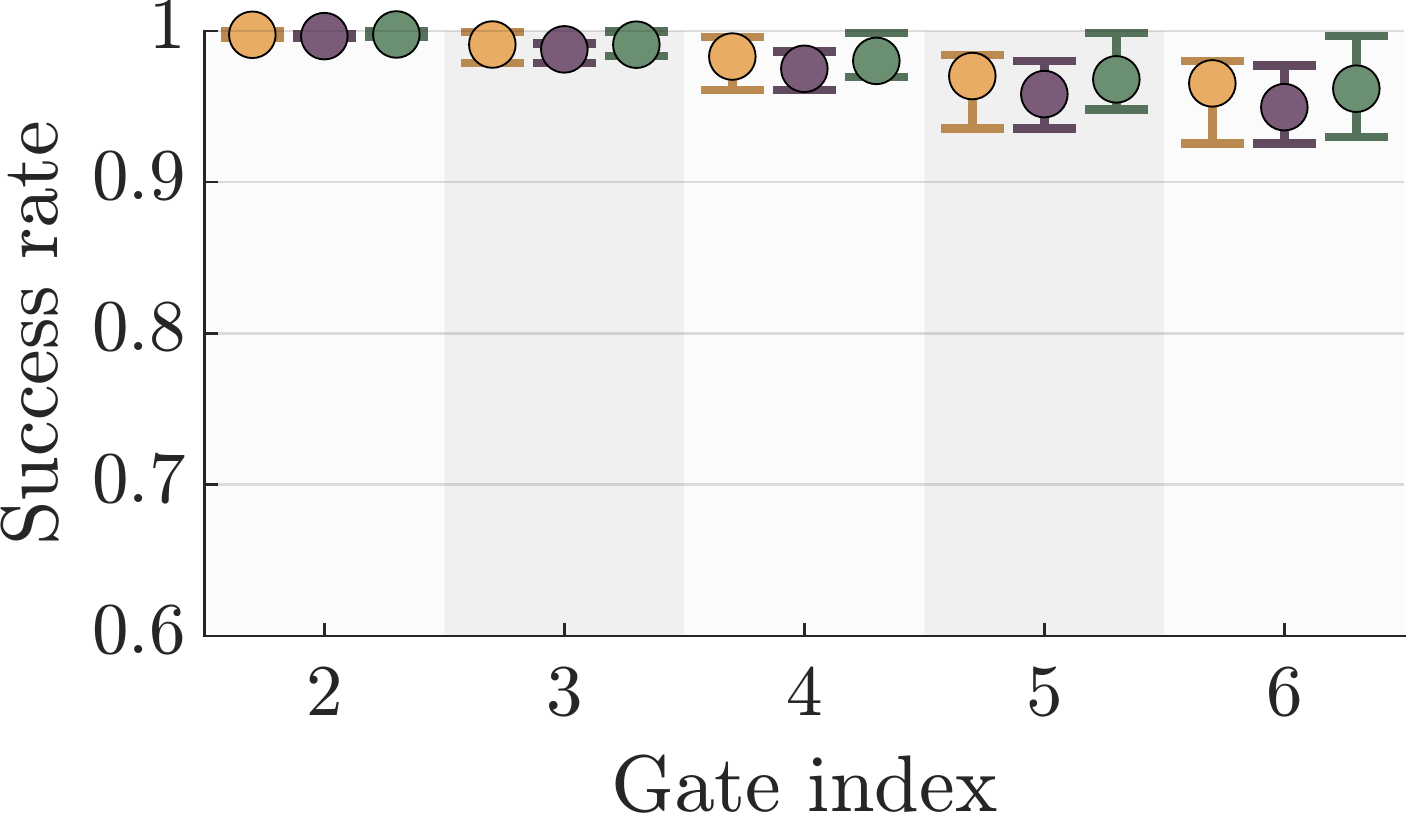}\label{fig:gate_sr_easy}}\\

% --- Row 2: Medium ---
\raisebox{1.2cm}{\rotatebox{90}{\small Medium}} &
\subfloat[]{\includegraphics[width=0.225\textwidth]{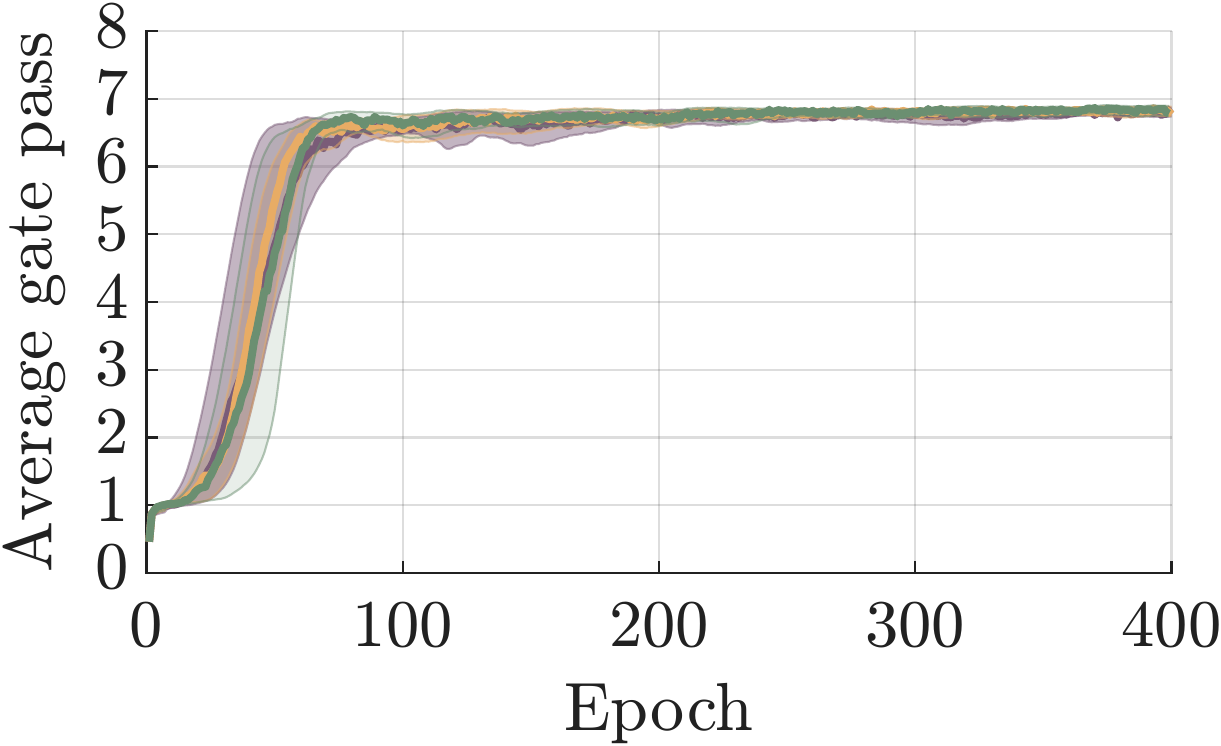}\label{fig:medium_gate_passed}} &
\subfloat[]{\includegraphics[width=0.23\textwidth]{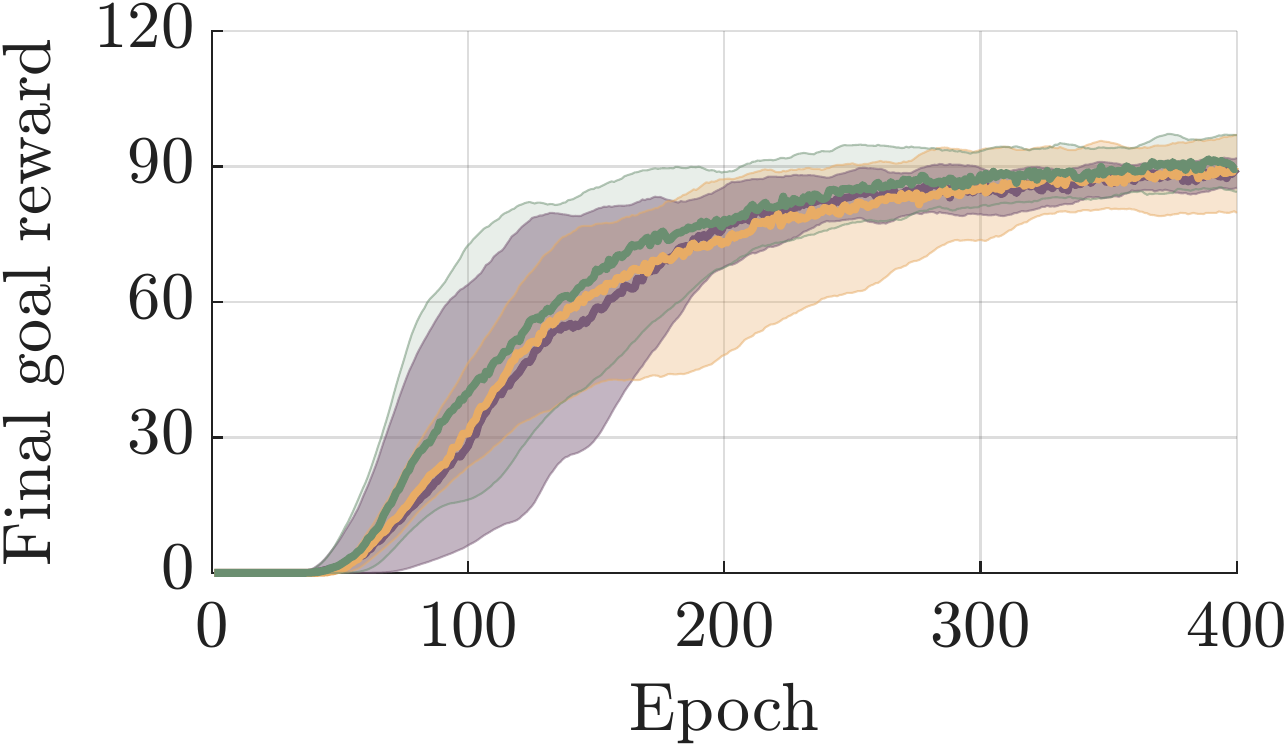}\label{fig:medium_goal_hover}} &
\subfloat[]{\includegraphics[width=0.2225\textwidth]{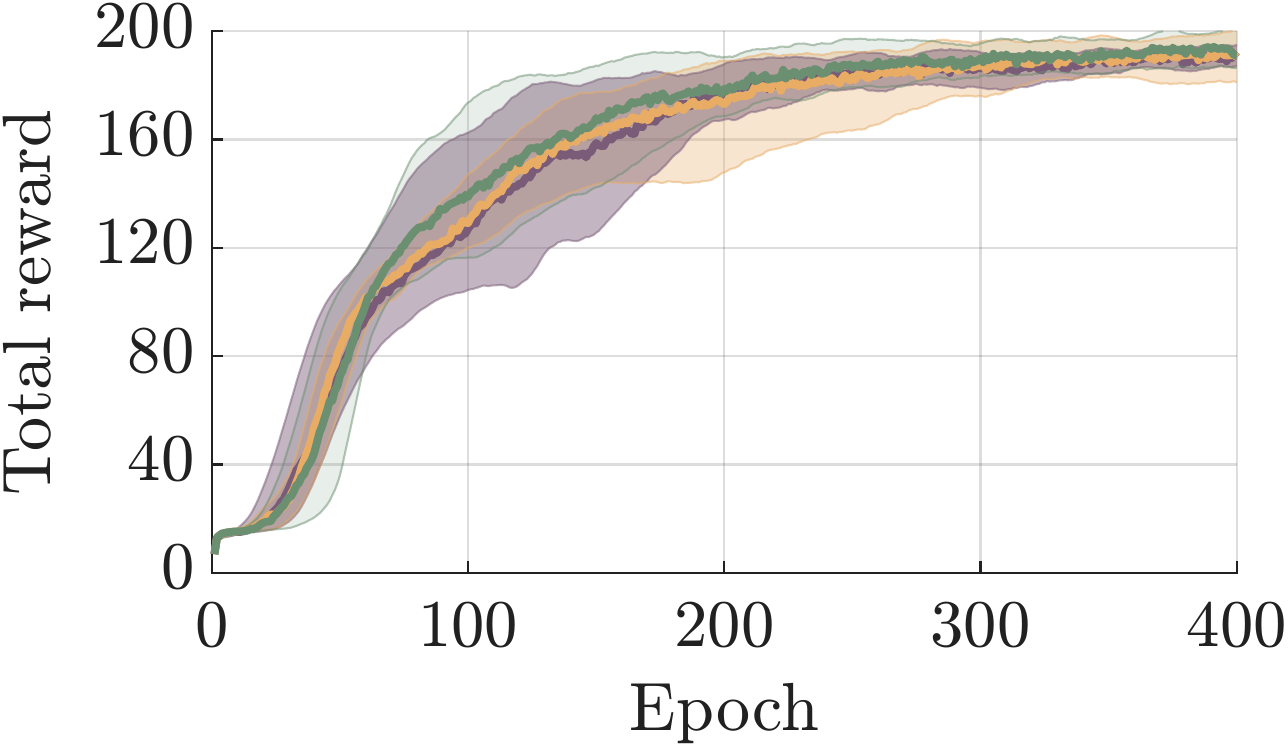}\label{fig:medium_total_reward}} &
\subfloat[]{\includegraphics[width=0.22\textwidth]{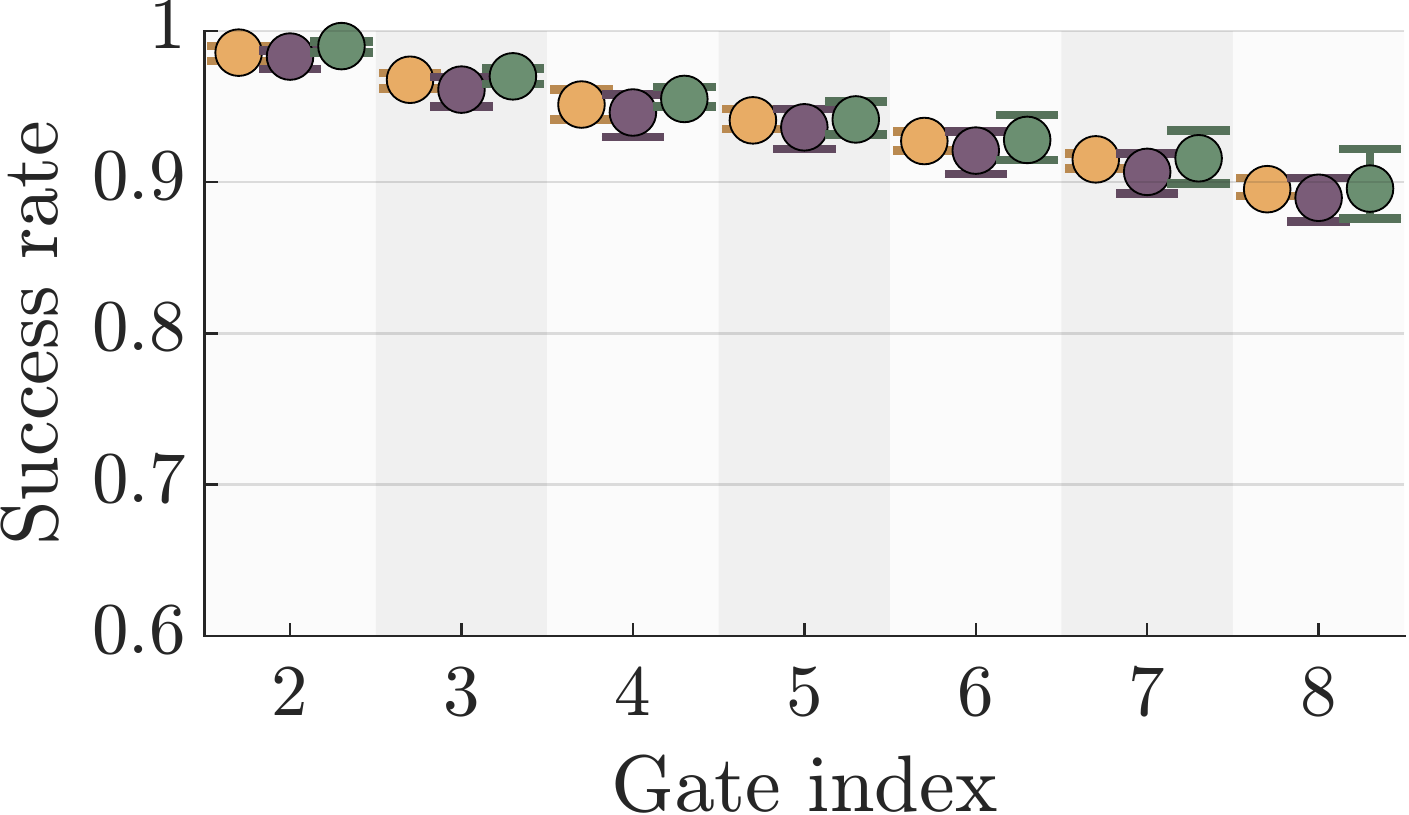}\label{fig:gate_sr_medium}}\\

% --- Row 3: Hard ---
\raisebox{1.2cm}{\rotatebox{90}{\small Hard}} &
\subfloat[]{\includegraphics[width=0.225\textwidth]{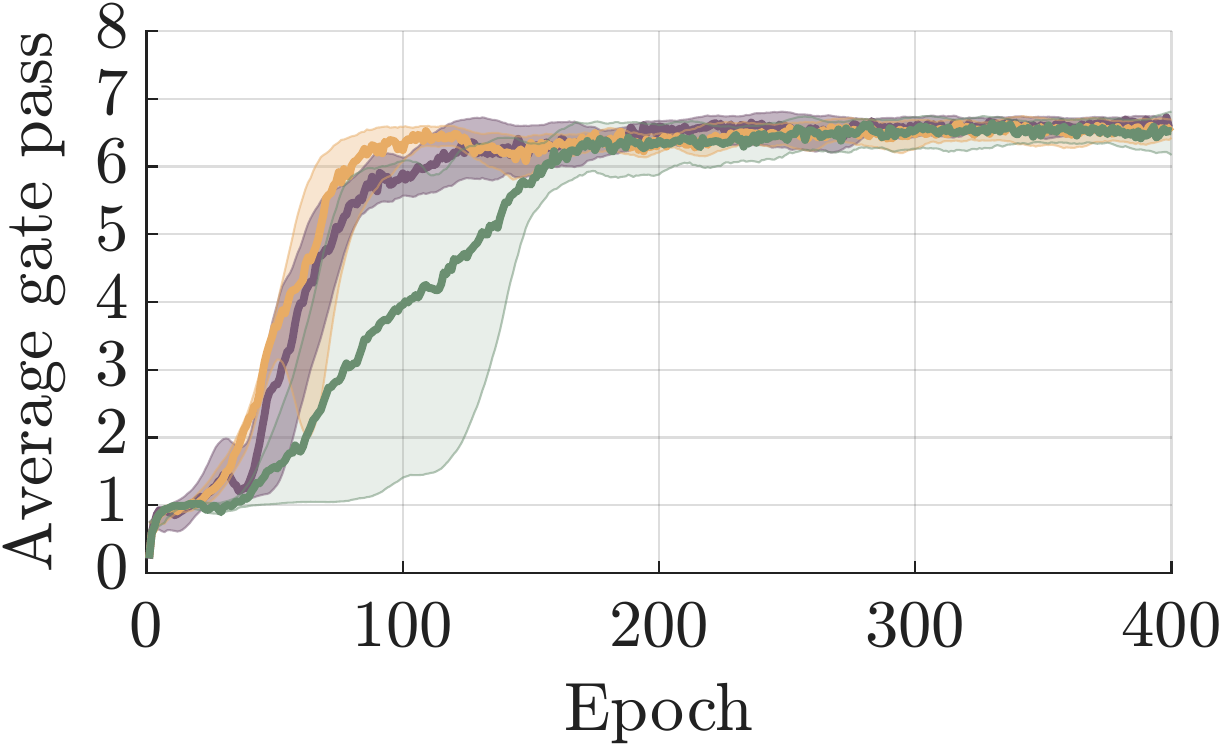}\label{fig:hard_gate_passed}} &
\subfloat[]{\includegraphics[width=0.23\textwidth]{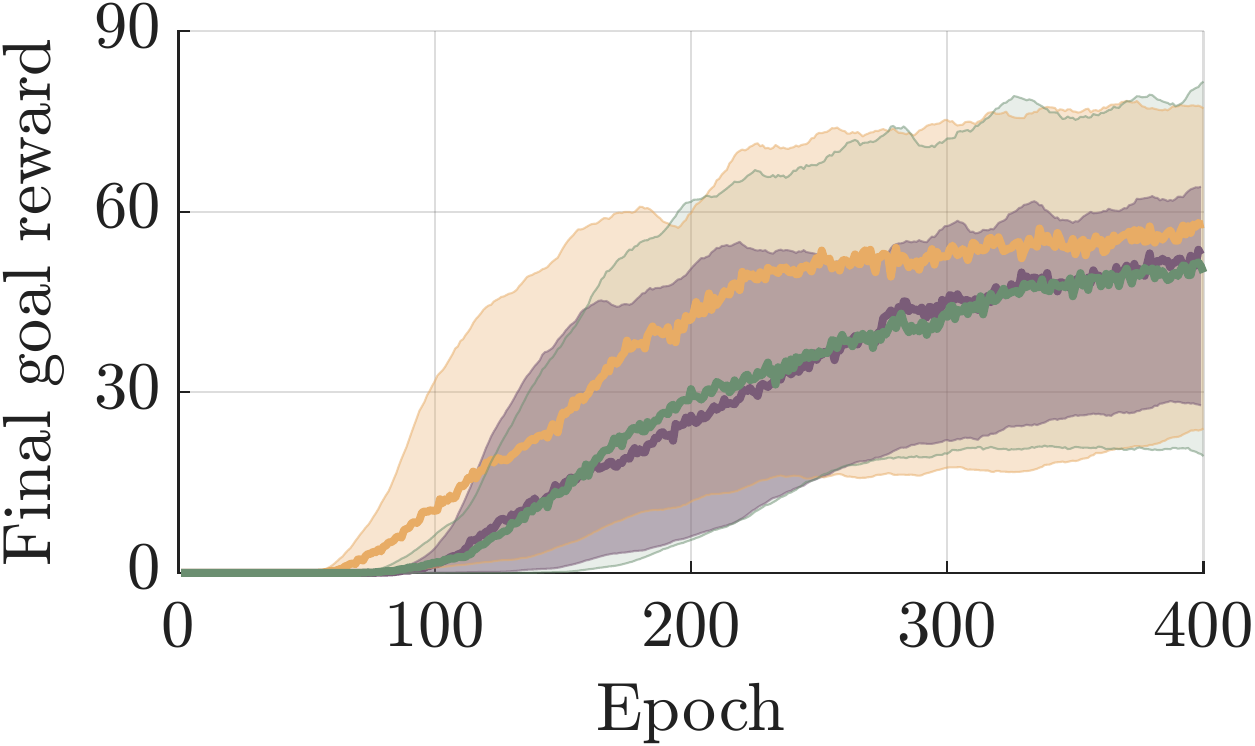}\label{fig:hard_goal_hover}} &
\subfloat[]{\includegraphics[width=0.23\textwidth]{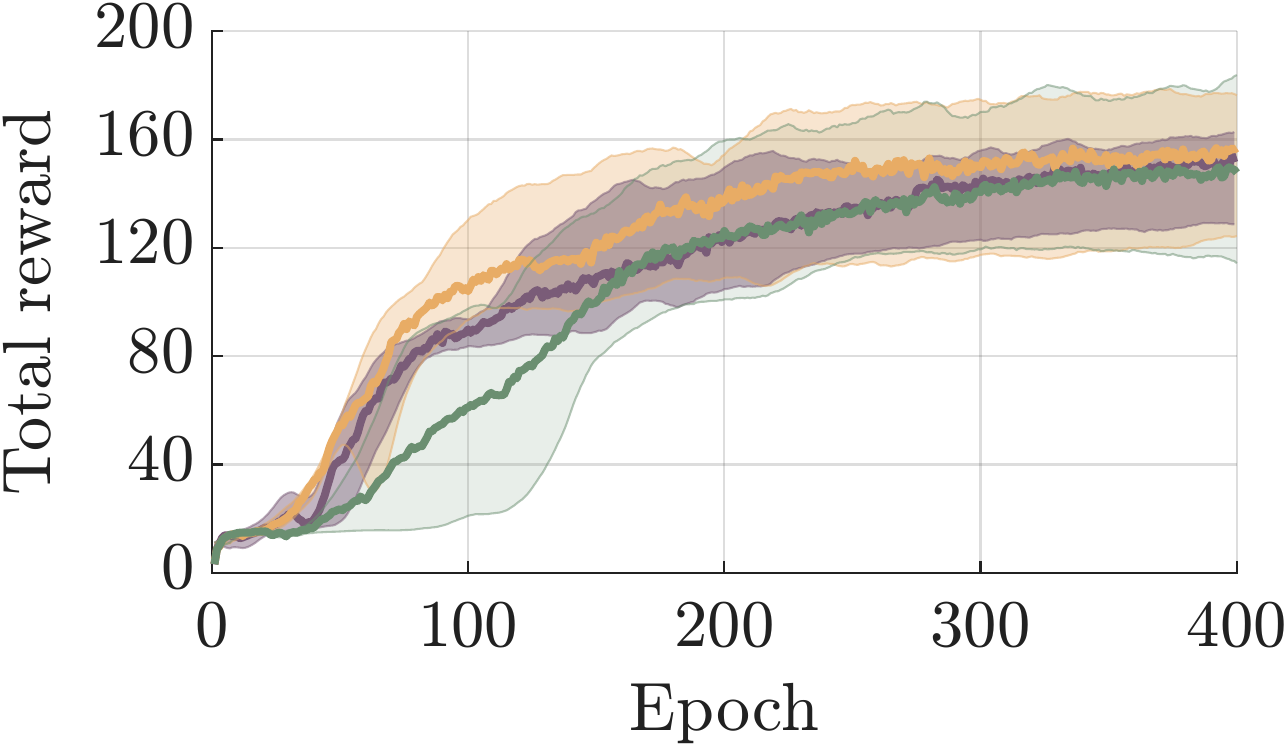}\label{fig:hard_total_reward}} &
\subfloat[]{\includegraphics[width=0.22\textwidth]{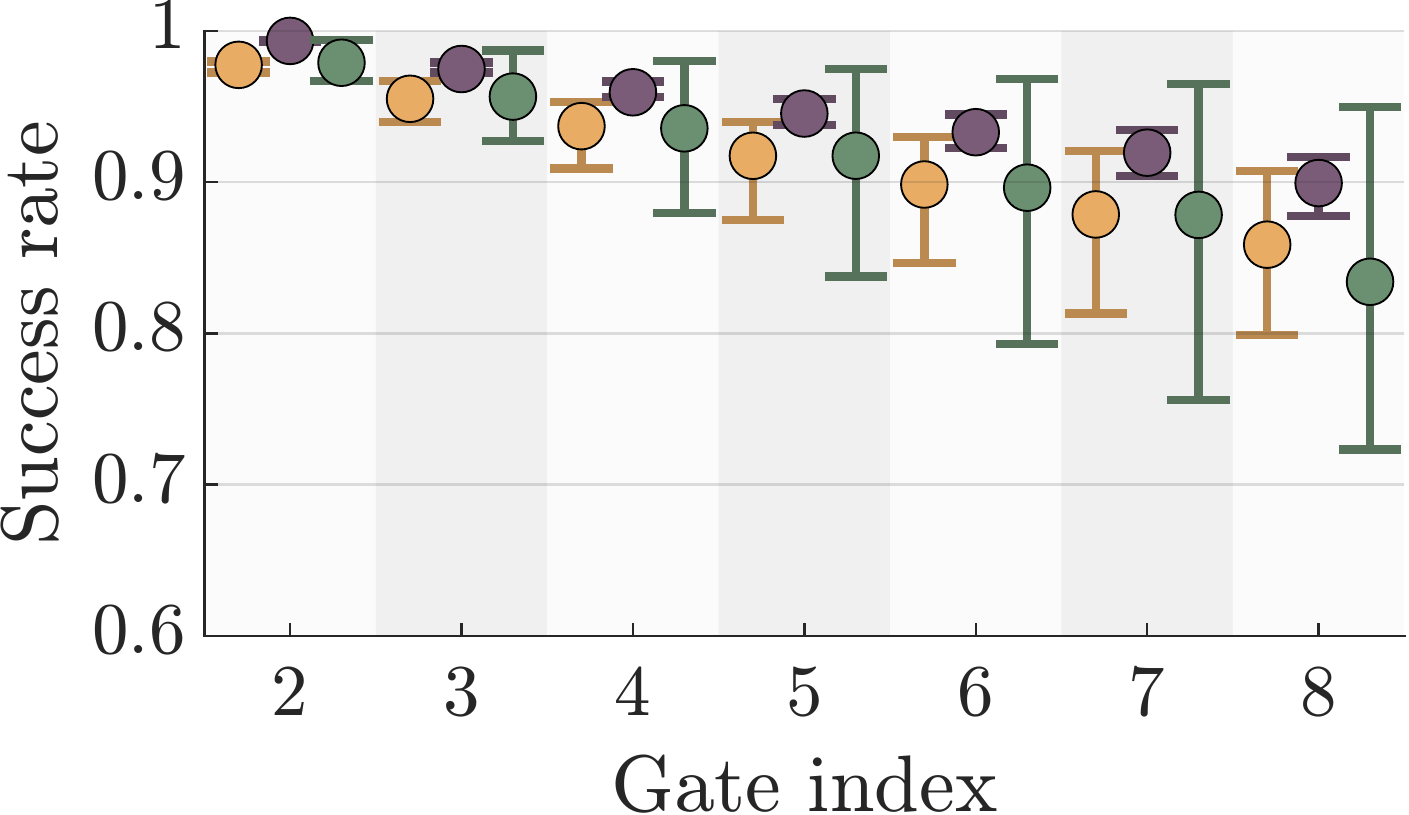}\label{fig:gate_sr_hard}}\\
\end{tabular}

% ================= LEGEND =================
\vspace{0.2em}
{\scriptsize
\textcolor[HTML]{6B8F71}{\rule{1.5em}{3pt}}\hspace{0.5em}non-fuzzy\hspace{0.6em}
\textcolor[HTML]{7A5C78}{\rule{1.5em}{3pt}}\hspace{0.5em}Mamdani\hspace{0.6em}
\textcolor[HTML]{E8AC65}{\rule{1.5em}{3pt}}\hspace{0.5em}Sugeno\hspace{0.6em}
}

\caption{
Training and evaluation results for the Easy, Medium, and Hard scenarios. The left three columns (a–c, e–g, i–k) report training performance, averaged over five random seeds. Thick solid lines denote the mean performance across seeds, while the semi-transparent shaded regions indicate variability across seeds.
The rightmost column (d, h, l) presents evaluation results, where each trained policy is tested over $5000$ evaluations. Circular markers denote the mean gate-passing success rate at each gate index, while vertical lines indicate the minimum–maximum range observed across all seeds. Success rates are normalized between $0$ and $1$, representing the probability of successfully passing each gate.  Results are shown for three reward configurations: non-fuzzy (green), Mamdani (purple), and Sugeno (yellow).
}
\label{fig:reward_breakdown_all}
\end{figure*}

All experiments are carried out using an actor–critic framework (PPO) with continuous action spaces, trained in IsaacLab~\cite{mittal2025isaaclab} with $2048$ parallel simulation environments and a horizon length of $128$. The policy and value networks share a multilayer perceptron architecture with hidden layers $[128, 128]$. Training employs an adaptive learning rate with an initial value of $4 \times 10^{-4}$, clipping $\epsilon = 0.2$, and entropy regularization with coefficient $0.01$. Policies are trained for up to $400$ epochs. All experiments are executed on a system equipped with an Intel® Core™ i7-14700K CPU and a NVIDIA GeForce RTX 4080 GPU.

\vspace{-0.2cm}
\section{Results}
\label{sec:results}
We evaluate the proposed fuzzy reward-shaping strategy on three scenarios of increasing difficulty (Easy, Medium, and Hard). For each scenario, we report training curves and evaluate the trained policies in the corresponding environment under a unified protocol. The protocol is summarized as follows:
\begin{itemize}[leftmargin=*]
    \item Three reward formulations—non-fuzzy PFBRS, Sugeno fuzzy-based reward shaping, and Mamdani fuzzy-based reward shaping—are evaluated across three scenarios (Easy, Medium, and Hard).
    
    \item All reward formulations described in Section~\ref{sec:experiments} are applied without any environment-specific re-tuning. In particular, fuzzy logic-based velocity--distance reward $r^{\mathrm{vd}}$ is kept unchanged across all scenarios and difficulty levels, ensuring that observed performance differences are not influenced by scenario-specific tuning.
    
    \item For each scenario and reward formulation, training is repeated with multiple random seeds\footnote{Seeds used: \{5, 8, 16, 32, 36\}}, and results are reported as the mean and variance across seeds. Figure~\ref{fig:reward_breakdown_all} decomposes the total reward into its components—gate-pass reward (Figs.~\ref{fig:easy_gate_passed}, \ref{fig:medium_gate_passed}, \ref{fig:hard_gate_passed}), hovering reward (Figs.~\ref{fig:easy_goal_hover}, \ref{fig:medium_goal_hover}, \ref{fig:hard_goal_hover}), and total reward (Figs.~\ref{fig:easy_total_reward}, \ref{fig:medium_total_reward}, \ref{fig:hard_total_reward})—during training. For evaluation, we execute each trained policy and report gate-passing performance as the number (and per-gate rate) of successfully traversed gates (Figs.~\ref{fig:gate_sr_easy}, \ref{fig:gate_sr_medium}, \ref{fig:gate_sr_hard}).
\end{itemize}
% Figure~\ref{fig:reward_breakdown_all} presents the breakdown of reward components for the Easy, Medium, and Hard scenarios, averaged over $7$ random seeds.

\subsection{Easy scenario}

In the Easy scenario, all reward configurations exhibit rapid convergence, as illustrated in Fig.~\ref{fig:easy_gate_passed}--\ref{fig:easy_total_reward}. As shown in Fig.~\ref{fig:easy_gate_passed}, gate-passing performance converges within approximately $50$ training epochs, with agents successfully traversing approximately $5$ out of $6$ gates on average across all seeds. This rapid convergence can be attributed to the reduced environmental stochasticity inherent in this scenario, including uniform gate heights and larger inter-gate spacing. Under these conditions, fuzzy-based reward formulations are observed to converge slightly earlier than their non-fuzzy counterparts for this metric; however, differences between reward formulations are not clearly distinguishable through the gate-passing metric alone. As shown in Fig.~\ref{fig:easy_goal_hover}, the final goal-hovering reward differs only marginally across the non-fuzzy, Mamdani, and Sugeno formulations. By the end of training, all three achieve comparable mean reward and variance, and the total accumulated reward in Fig.~\ref{fig:easy_total_reward} is similarly aligned, indicating equivalent performance in the Easy scenario.

Evaluation results for gate traversal success rate shown in Fig.~\ref{fig:gate_sr_easy}, further indicates that all three reward formulations achieve very similar final-gate success rates in the Easy scenario. On average, the probability of passing the final gate exceeds $95\%$ across methods, reflecting the overall simplicity of the task. While the non-fuzzy formulation reaches $100\%$ success in some individual test runs, its average performance remains closely aligned with the Sugeno-based formulation, with only marginal differences observed across seeds.

\subsection{Medium scenario}

%As shown in Fig.~\ref{fig:medium_gate_passed}, agents trained with all reward formulations consistently learn to traverse the gate sequence within the first 60 steps of each episode.

In the Medium scenario, all reward formulations exhibit stable convergence across all seeds, with learning dynamics closely resembling those observed in the Easy case, albeit with increased variability.
The gate-pass reward in Fig.~\ref{fig:medium_gate_passed} converges within approximately 60 training epochs, with agents traversing about 7 out of 8 gates on average across all seeds. Although all methods converge, the final goal-hovering reward in Fig.~\ref{fig:medium_goal_hover} is lower than in the Easy scenario, consistent with the increased task difficulty and the reduced time available for hovering at the final goal. Figure~\ref{fig:medium_total_reward} reports the total reward, showing convergence for all methods within approximately 300 training epochs.
The gate traversal success rate, as shown in Fig.~\ref{fig:gate_sr_medium}, is used for evaluation in this scenario and indicates that all reward formulations achieve high success rates. The probability of passing the final gate exceeds 90\% for all three methods, demonstrating robust task completion. Among them, the non-fuzzy formulation attains a slightly higher final success rate, although the overall differences remain marginal.

%Total accumulated rewards converge to similar final values across methods, with the non-fuzzy formulation achieving slightly higher rewards. However, increased variance prevents a clear separation between reward formulations in the Medium scenario.

\subsection{Hard scenario}
In the Hard scenario, increased yaw perturbations and narrower gate geometries lead to clearer performance differences between reward formulations. 
While all methods converge across all seeds, fuzzy-based rewards achieve faster and more consistent gate-passing convergence, stabilizing after approximately $100$ training epochs, compared to around $150$ epochs for the non-fuzzy formulation (Fig.~\ref{fig:hard_gate_passed}). All methods converge to an average gate-passing value above $6$, although the non-fuzzy approach exhibits higher variability across seeds during early training.

Despite achieving rapid gate traversal after convergence, the non-fuzzy formulation progresses more slowly in the final goal hovering task than the Mamdani and Sugeno formulations (Fig.~\ref{fig:hard_goal_hover}). Among all methods, the Sugeno-based reward demonstrates the most stable learning behavior and achieves the highest total accumulated reward in the Hard scenario (Fig.~\ref{fig:hard_total_reward}).

The gate traversal success rate for the Hard scenario is illustrated in Fig.~\ref{fig:gate_sr_hard}. The Mamdani formulation achieves the highest average success rate, exceeding $90\%$, with the lowest variance across seeds. Although the non-fuzzy formulation reaches up to $95\%$ success in some runs, its performance fluctuates substantially across seeds. Overall, these results indicate that fuzzy-based reward formulations provide improved robustness under increased task difficulty.

\section{Conclusion}

This paper introduces \algname, a fuzzy logic--based reward shaping method that incorporates interpretable, human-inspired heuristics into RL. By encoding domain knowledge through linguistic rules, the proposed approach adaptively modulates velocity--distance and task-related reward components based on the agent’s state, providing an alternative to fixed, hand-crafted reward formulations. The method is evaluated on autonomous drone racing scenarios of increasing difficulty using three reward formulations: non-fuzzy, Mamdani-based, and Sugeno-based fuzzy inference. In simpler environments, all three methods achieve comparable performance, with only marginal differences observed and, in some cases, slightly better results obtained by the non-fuzzy formulation. As task complexity increases, fuzzy-based reward formulations tend to converge faster and exhibit more stable behavior across training seeds, resulting in more consistent gate-passing performance. These findings highlight the potential of fuzzy logic--based reward shaping as an effective design choice in complex and variable settings, while maintaining competitive performance in simpler scenarios.

Future work will focus on real-time evaluation to assess robustness beyond simulation and analyze sim-to-real transfer performance. We will also investigate unifying different reward formulations into a compact fuzzy framework, along with learning-based adaptation of membership functions and rule sets during training.
\label{sec:conclusion}
\bibliographystyle{IEEEtran}
\bibliography{References}

@misc{schulman2017proximal,
      title={Proximal Policy Optimization Algorithms}, 
      author={John Schulman and Filip Wolski and Prafulla Dhariwal and Alec Radford and Oleg Klimov},
      year={2017},
      eprint={1707.06347},
      archivePrefix={arXiv},
      primaryClass={cs.LG},
      url={https://arxiv.org/abs/1707.06347}, 
}

@ARTICLE{11152316,
  author={Dang, Van Huyen and Redder, Adrian and Pham, Huy Xuan and Sarabakha, Andriy and Kayacan, Erdal},
  journal={IEEE Robotics and Automation Letters}, 
  title={VDS-Nav: Volumetric Depth-Based Safe Navigation for Aerial Robots–Bridging the Sim-to-Real Gap}, 
  year={2025},
  volume={10},
  number={10},
  pages={11038-11045},
  doi={10.1109/LRA.2025.3606806}}

@INPROCEEDINGS{10610910,
  author={Sayar, Erdi and Bing, Zhenshan and D’Eramo, Carlo and Oguz, Ozgur S. and Knoll, Alois},
  booktitle={2024 IEEE International Conference on Robotics and Automation (ICRA)}, 
  title={Contact Energy Based Hindsight Experience Prioritization}, 
  year={2024},
  volume={},
  number={},
  pages={5434-5440},
  doi={10.1109/ICRA57147.2024.10610910}}

@INPROCEEDINGS{11007627,
  author={Ugurlu, Halil Ibrahim and Redder, Adrian and Kayacan, Erdal},
  booktitle={2025 IEEE Symposium on Computational Intelligence on Engineering/Cyber Physical Systems (CIES)}, 
  title={Lyapunov-Inspired Deep Reinforcement Learning for Robot Navigation in Obstacle Environments}, 
  year={2025},
  volume={},
  number={},
  pages={1-8},
  doi={10.1109/CIES64955.2025.11007627}}

@inproceedings{10.1145/3638529.3654045,
author = {Sayar, Erdi and Iacca, Giovanni and Knoll, Alois},
title = {Multi-Objective Evolutionary Hindsight Experience Replay for Robot Manipulation Tasks},
year = {2024},
isbn = {9798400704949},
publisher = {Association for Computing Machinery},
address = {New York, NY, USA},
url = {https://doi.org/10.1145/3638529.3654045},
doi = {10.1145/3638529.3654045},
booktitle = {Proceedings of the Genetic and Evolutionary Computation Conference},
pages = {403–411},
numpages = {9},
location = {Melbourne, VIC, Australia},
series = {GECCO '24}
}

@ARTICLE{9894655,
  author={Pham, Huy Xuan and Sarabakha, Andriy and Odnoshyvkin, Mykola and Kayacan, Erdal},
  journal={IEEE Robotics and Automation Letters}, 
  title={PencilNet: Zero-Shot Sim-to-Real Transfer Learning for Robust Gate Perception in Autonomous Drone Racing}, 
  year={2022},
  volume={7},
  number={4},
  pages={11847-11854},
  doi={10.1109/LRA.2022.3207545}}

@INPROCEEDINGS{10649903,
  author={Qiao, Zhongzheng and Pham, Xuan Huy and Ramasamy, Savitha and Jiang, Xudong and Kayacan, Erdal and Sarabakha, Andriy},
  booktitle={2024 International Joint Conference on Neural Networks (IJCNN)}, 
  title={Continual Learning for Robust Gate Detection under Dynamic Lighting in Autonomous Drone Racing}, 
  year={2024},
  volume={},
  number={},
  pages={1-8},
  doi={10.1109/IJCNN60899.2024.10649903}}

@Article{robotics11050109,
AUTHOR = {Ugurlu, Halil Ibrahim and Pham, Xuan Huy and Kayacan, Erdal},
TITLE = {Sim-to-Real Deep Reinforcement Learning for Safe End-to-End Planning of Aerial Robots},
JOURNAL = {Robotics},
VOLUME = {11},
YEAR = {2022},
NUMBER = {5},
ARTICLE-NUMBER = {109},
URL = {https://www.mdpi.com/2218-6581/11/5/109},
ISSN = {2218-6581},
DOI = {10.3390/robotics11050109}
}

@INPROCEEDINGS{9838538,
  author={Andersen, Kristoffer Fogh and Pham, Huy Xuan and Ugurlu, Halil Ibrahim and Kayacan, Erdal},
  booktitle={2022 European Control Conference (ECC)}, 
  title={Event-based Navigation for Autonomous Drone Racing with Sparse Gated Recurrent Network}, 
  year={2022},
  volume={},
  number={},
  pages={1342-1348},
  doi={10.23919/ECC55457.2022.9838538}}

@INPROCEEDINGS{9636207,
  author={Pham, Huy Xuan and Bozcan, Ilker and Sarabakha, Andriy and Haddadin, Sami and Kayacan, Erdal},
  booktitle={2021 IEEE/RSJ International Conference on Intelligent Robots and Systems (IROS)}, 
  title={GateNet: An Efficient Deep Neural Network Architecture for Gate Perception Using Fish-Eye Camera in Autonomous Drone Racing}, 
  year={2021},
  volume={},
  number={},
  pages={4176-4183},
  doi={10.1109/IROS51168.2021.9636207}}

@INPROCEEDINGS{9206943,
  author={Morales, Théo and Sarabakha, Andriy and Kayacan, Erdal},
  booktitle={2020 International Joint Conference on Neural Networks (IJCNN)}, 
  title={Image Generation for Efficient Neural Network Training in Autonomous Drone Racing}, 
  year={2020},
  volume={},
  number={},
  pages={1-8},
  doi={10.1109/IJCNN48605.2020.9206943}}

@incollection{PHAM2022371,
title = {Chapter 15 - Deep learning for vision-based navigation in autonomous drone racing},
editor = {Alexandros Iosifidis and Anastasios Tefas},
booktitle = {Deep Learning for Robot Perception and Cognition},
publisher = {Academic Press},
pages = {371-406},
year = {2022},
isbn = {978-0-323-85787-1},
doi = {https://doi.org/10.1016/B978-0-32-385787-1.00020-8},
author = {Huy Xuan Pham and Halil Ibrahim Ugurlu and Jonas {Le Fevre} and Deniz Bardakci and Erdal Kayacan},}

@ARTICLE{faessler2018differential,
  author={Faessler, Matthias and Franchi, Antonio and Scaramuzza, Davide},
  journal={IEEE Robotics and Automation Letters},
  title={Differential Flatness of Quadrotor Dynamics Subject to Rotor Drag for Accurate Tracking of High-Speed Trajectories},
  year={2018},
  volume={3},
  number={2},
  pages={620-626},
  doi={10.1109/LRA.2017.2776353}}

@ARTICLE{lyapunov_fuzzy,
  author={Chen, Ming and Lam, Hak Keung and Shi, Qian and Xiao, Bo},
  journal={IEEE Transactions on Circuits and Systems II: Express Briefs}, 
  title={Reinforcement Learning-Based Control of Nonlinear Systems Using Lyapunov Stability Concept and Fuzzy Reward Scheme}, 
  year={2020},
  volume={67},
  number={10},
  pages={2059-2063},
  doi={10.1109/TCSII.2019.2947682}}

@Article{Bingolsafe,
AUTHOR = {Bingol, Mustafa Can},
TITLE = {A Safe Navigation Algorithm for Differential-Drive Mobile Robots by Using Fuzzy Logic Reward Function-Based Deep Reinforcement Learning},
JOURNAL = {Electronics},
VOLUME = {14},
YEAR = {2025},
NUMBER = {8},
ARTICLE-NUMBER = {1593},
ISSN = {2079-9292},
DOI = {10.3390/electronics14081593}
}

@ARTICLE{7580570,
  author={Kayacan, Erdal and Maslim, Reinaldo},
  journal={IEEE/ASME Transactions on Mechatronics}, 
  title={Type-2 Fuzzy Logic Trajectory Tracking Control of Quadrotor VTOL Aircraft With Elliptic Membership Functions}, 
  year={2017},
  volume={22},
  number={1},
  pages={339-348},
  doi={10.1109/TMECH.2016.2614672}}

@misc{yu2025master,
      title={Mastering Diverse, Unknown, and Cluttered Tracks for Robust Vision-Based Drone Racing}, 
      author={Feng Yu and Yu Hu and Yang Su and Yang Deng and Linzuo Zhang and Danping Zou},
      year={2025},
      eprint={2512.09571},
      archivePrefix={arXiv},
      primaryClass={cs.RO},
}

@misc{wang2026,
      title={Environment as Policy: Learning to Race in Unseen Tracks}, 
      author={Hongze Wang and Jiaxu Xing and Nico Messikommer and Davide Scaramuzza},
      year={2026},
      eprint={2410.22308},
      archivePrefix={arXiv},
      primaryClass={cs.RO},
}

@inproceedings{ng1999policy,
  title={Policy invariance under reward transformations: Theory and application to reward shaping},
  author={Ng, Andrew Y and Harada, Daishi and Russell, Stuart},
  booktitle={Icml},
  volume={99},
  pages={278--287},
  year={1999},
  organization={Citeseer}
}

@article{mittal2025isaaclab,
  title={Isaac lab: A gpu-accelerated simulation framework for multi-modal robot learning},
  author={Mittal, Mayank and Roth, Pascal and Tigue, James and Richard, Antoine and Zhang, Octi and Du, Peter and Serrano-Munoz, Antonio and Yao, Xinjie and Zurbr{\"u}gg, Ren{\'e} and Rudin, Nikita and others},
  journal={arXiv preprint arXiv:2511.04831},
  year={2025}
}

@incollection{puterman1990markov,
title = {Chapter 8 Markov decision processes},
series = {Handbooks in Operations Research and Management Science},
publisher = {Elsevier},
volume = {2},
pages = {331-434},
year = {1990},
booktitle = {Stochastic Models},
issn = {0927-0507},
doi = {https://doi.org/10.1016/S0927-0507(05)80172-0},
url = {https://www.sciencedirect.com/science/article/pii/S0927050705801720},
author = {Martin L. Puterman},
}

@book{sugeno1985fuzzy,
author = {Sugeno, Michio},
title = {Industrial Applications of Fuzzy Control},
year = {1985},
isbn = {0444878297},
publisher = {Elsevier Science Inc.},
address = {USA}
}

@article{mamdani1974application,
author = {E.H. Mamdani },
title = {Application of fuzzy algorithms for control of simple dynamic plant},
journal = {Proceedings of the Institution of Electrical Engineers},
volume = {121},
issue = {12},
pages = {1585-1588},
year = {1974},
doi = {10.1049/piee.1974.0328},

URL = {https://digital-library.theiet.org/doi/abs/10.1049/piee.1974.0328},
eprint = {https://digital-library.theiet.org/doi/pdf/10.1049/piee.1974.0328}
}

@article{CAMCI20181,
title = {An aerial robot for rice farm quality inspection with type-2 fuzzy neural networks tuned by particle swarm optimization-sliding mode control hybrid algorithm},
journal = {Swarm and Evolutionary Computation},
volume = {41},
pages = {1-8},
year = {2018},
issn = {2210-6502},
doi = {https://doi.org/10.1016/j.swevo.2017.10.003},
url = {https://www.sciencedirect.com/science/article/pii/S2210650217302407},
author = {Efe Camci and Devesh Raju Kripalani and Linlu Ma and Erdal Kayacan and Mojtaba Ahmadieh Khanesar},
}

@ARTICLE{8304792,
  author={Fu, Changhong and Sarabakha, Andriy and Kayacan, Erdal and Wagner, Christian and John, Robert and Garibaldi, Jonathan M.},
  journal={IEEE/ASME Transactions on Mechatronics}, 
  title={Input Uncertainty Sensitivity Enhanced Nonsingleton Fuzzy Logic Controllers for Long-Term Navigation of Quadrotor UAVs}, 
  year={2018},
  volume={23},
  number={2},
  pages={725-734},
  doi={10.1109/TMECH.2018.2810947}}

@ARTICLE{8809217,
  author={Sarabakha, Andriy and Kayacan, Erdal},
  journal={IEEE Transactions on Fuzzy Systems}, 
  title={Online Deep Fuzzy Learning for Control of Nonlinear Systems Using Expert Knowledge}, 
  year={2020},
  volume={28},
  number={7},
  pages={1492-1503},
  doi={10.1109/TFUZZ.2019.2936787}}

@article{SARABAKHA2019105495,
title = {Intuit before tuning: Type-1 and type-2 fuzzy logic controllers},
journal = {Applied Soft Computing},
volume = {81},
pages = {105495},
year = {2019},
issn = {1568-4946},
doi = {https://doi.org/10.1016/j.asoc.2019.105495},
url = {https://www.sciencedirect.com/science/article/pii/S1568494619302650},
author = {Andriy Sarabakha and Changhong Fu and Erdal Kayacan},
}

@INPROCEEDINGS{7737744,
  author={Camci, Efe and Kayacan, Erdal},
  booktitle={2016 IEEE International Conference on Fuzzy Systems (FUZZ-IEEE)}, 
  title={Game of drones: UAV pursuit-evasion game with type-2 fuzzy logic controllers tuned by reinforcement learning}, 
  year={2016},
  volume={},
  number={},
  pages={618-625},
  doi={10.1109/FUZZ-IEEE.2016.7737744}}
\end{document}